\pdfoutput=1

\documentclass[11pt]{article}

\usepackage{EMNLP2022}

\usepackage{times}
\usepackage{latexsym}

\usepackage[T1]{fontenc}

\usepackage[utf8]{inputenc}

\usepackage{microtype}

\usepackage{inconsolata}

\usepackage{enumitem}
\usepackage{multirow}
\usepackage{graphicx}
\usepackage{tabularx}
\usepackage{booktabs}
\usepackage{amssymb}
\usepackage{amsmath}
\usepackage{amsfonts}
\usepackage{centernot}
\usepackage[ruled,vlined]{algorithm2e}

\SetCommentSty{mycommfont}

\newcommand{\good}{\textcolor{blue}}

\newcommand{\tess}{\texttt{TESS}\xspace}

\title{\tess: A Multi-intent Parser for Conversational Multi-Agent Systems with Decentralized Natural Language Understanding Models}

\author{
Burak Aksar$^1$\thanks{\,\,\, This work was done during Burak's internship at IBM Research during the summers of 2021 and 2022. The results, the choice of baselines, and the conclusions drawn are representative of that timeline.}
$\cdot$ Yara Rizk$^2$ $\cdot$ Tathagata Chakraborti$^2$ 
\\[0.75ex]
$^1$Boston University, $^2$IBM Research
\\[0.5ex]
$^1$Contact: \texttt{baksar@bu.edu}
}

\vspace{5mm}

\begin{document}
\maketitle
\begin{abstract}
Chatbots have become one of the main pathways for the delivery of business automation tools. Multi-agent systems offer a framework for designing chatbots at scale, making it easier to support complex conversations that span across multiple domains as well as enabling developers to maintain and expand their capabilities incrementally over time.  However, multi-agent systems complicate the natural language understanding (NLU) of user intents, especially when they rely on decentralized NLU models: some utterances (termed single intent) may invoke a single agent while others (termed multi-intent) may explicitly invoke multiple agents.  Without correctly parsing multi-intent inputs, decentralized NLU approaches will not achieve high prediction accuracy. In this paper, we propose an efficient parsing and orchestration pipeline algorithm to service multi-intent utterances from the user in the context of a multi-agent system. Our proposed approach achieved comparable performance to competitive deep learning models on three different datasets while being up to 48 times faster. 

\end{abstract}

\section{Introduction}

The backbone of any commercial venture or organization is its business process portfolio~\cite{weske2004advances}. A business process is a set of operations or activities that must be completed in a specific order to achieve a task \cite{dumas2013fundamentals}. Robotic process automation (RPA) is a low-cost technique for injecting automation into business operations in the age of digital transformation; RPA bots are ideal for tasks that are mundane, repetitive, and error-prone~\cite{ivanvcic2019robotic}.

However, most business users who would benefit from RPAs are not technically savvy enough to fully leverage their potential~\cite{jakob2018barriers}. This lack of accessibility limits a business user's capacity to monitor and adapt such solutions~\cite{gao2019automated}. Another limiting factor is that highly-regulated industries still require human-in-the-loop automation due to security and liability issues. As a result, RPA vendors have provided conversational interfaces to their bots to make them more accessible to business users. 

A digital assistant can be made up of several conversational software bots that automate specific tasks for a business process. It enables business users and domain experts who lack programming or software development abilities to design and communicate with their business process automation solutions using natural language. Digital assistants provide a natural language interface to invoke and interact with RPA bots \cite{anagnoste2021role} to lower the barrier to entry for process automation. RPA bots have been integrated with chatbot authoring tools like Dialogflow \cite{gajra2020automating} and in multi-agent conversational systems like \citet{rizk2020conversational}. As noted in the latter, in addition to accessibility, 
a multi-agent framework makes the system easily extensible for developers.

However, the complexity of semantic parsing and intent detection increases in chatbots implemented as multi-agent systems featuring decentralized natural language understanding (NLU) models. Natural language utterances may range from simple single-intent phrases that map to a single to more complex phrases that map to multiple agents. State-of-the-art semantic parsers, capable of understanding a wide range of complex phrases, generally rely on deep language models that are computationally demanding and not easily customizable to specialized domains \cite{xuan2020improving,einolghozati2019improving}. Businesses that leverage RPAs often do not have the compute power (GPU clusters, etc.) to fine-tune or run deep learning models \cite{bauer2020machine,schlogl2019artificial}. Furthermore, their specific domains may not have enough training data to fine-tune these models \cite{al2019application,rongali2022training}.

In this study, we introduce a computationally efficient decentralized semantic parser tailored for multi-intent phrases within multi-agent conversational systems. This parser utilizes a posterior orchestration algorithm coupled with decentralized natural language evaluators to ascertain the most probable parsing. Empirical evidence indicates that our method attains accuracy on par with deep learning algorithms yet requires substantially less computational power.

\begin{figure}[tb]
\centering
\includegraphics[width=\columnwidth]{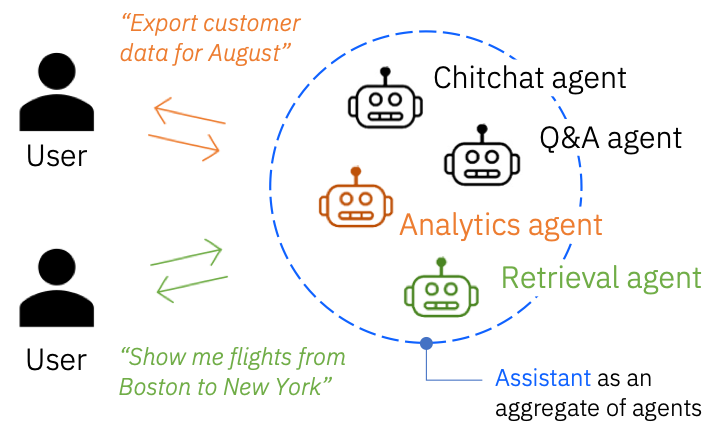}
\caption{
Scaling up conversational assistants using a distributed architecture. The user is talking to a singular assistant across multiple domains, and the conversation is handled internally by respective agents.
}
\label{fig:assistant_summary}
\end{figure}

\section{Conversational Systems at Scale}

While monolithic dialogue agents can be constructed using a combination of  intents and entities, such as in 
Watson Assistant\footnote{https://www.ibm.com/products/watson-assistant},  Dialogflow\footnote{https://cloud.google.com/dialogflow/docs/}, and Rasa\footnote{https://rasa.com/}, such approaches do not scale \cite{calhoun,martin,lior}  to conversations of sufficient complexity due to the developer needing to manually specify a complex dialogue tree as an (exponential) combination of those intents and entities. As a monolith, this approach is not only difficult in terms of coverage of all the cases to be covered by the assistant but also almost impossible to maintain and refactor over time once new improvements or additions are needed \cite{actions}.
As a result, there has been an increasing movement towards more dynamic approaches to the design of conversational agents using task-oriented declarative elements \cite{actions,salesforce,muise}.

\subsection{Scaling with an Aggregated Architecture}

The power of using dynamic composition solely depends on how reusable the individual elements in the system are. Of late, an architecture that has become increasingly powerful in this context is that of an assistant realized as an aggregate of elements called agents (or skills) \cite{rizk2020conversational}. These form the units of automation, and the capabilities of the assistant are realized as an aggregate of these individual units. While agents may or may not be conversational, in this paper, we will consider only those directly invoked by natural language utterances (as it relates to the multi-intent parsing task addressed in this work).

Examples of such agents include chit-chat and Q\&A, information retrieval,
data analytics, and so on. Prime examples of this include the skill catalog of Amazon  Alexa\footnote{https://developer.amazon.com/alexa/alexa-skills-kit} in the space of commercial assistants and similar catalogs for assistants for enterprise applications from 
Automation Anywhere\footnote{https://botstore.automationanywhere.com}, UiPath\footnote{https://marketplace.uipath.com/listings}, etc. Such an assistant is conceptualized in Figure ~\ref{fig:assistant_summary}, where the user is having a multi-domain conversation with the same
assistant (internally handled by respective agents).

Using a distributed architecture, it is possible to achieve an exponential scale-up from the complexity of specification to the sophistication of conversational paths supported by the assistant (or, conversely, an exponential reduction of the size of specification for the same complexity of the assistant). Past works have demonstrated this by using automated composition techniques \cite{aggr,d3ba}.

Formally, we can describe such an assistant $\mathbb{A}$ as a functional mapping from an event $\mathcal{E}$ (e.g., natural language utterance) and a set of agents $\Phi$ to a new event (e.g., a function call or a natural
language response to the user). The nature of this mapping describes the properties of the assistant. 
\begin{equation}
\label{eqn:assistant}
\mathbb{A} : \mathcal{E} \times \Phi \mapsto \mathcal{E}
\end{equation}

\subsection{The Curse of Scaling}

This decentralized approach allows developers to build and maintain their individual agents without having to worry about the rest of the capabilities of the assistant that is going to house them. However, the immediate impact of this setup is that the assistant has lost control and influence over the capabilities of its constituent elements or agents.

\subsubsection{Decentralized Knowledge}
\label{subsec:evaluation}

Since these agents are developed and maintained independently, the assistant by itself can no longer determine 
which agents can handle a user query. This problem is exacerbated by the fact that it is impossible to predict the correct way to parse a user utterance without knowledge of what each individual agent does. Consider, for example, the following utterance: 

\begin{quote}
\em
$S_1$: ``Book me a hotel and flight from Boston to New York for this weekend.''
\end{quote}

This could be a single-intent utterance if there is an agent in the orchestration that can book both flights and hotels but is a multi-intent utterance if there are agents that deal with flights and hotels separately, in which case, both agents must be executed in response to this request. As a first step, we can delegate that determination to the agents themselves, for them to self-report their relevance to the user query, thereby allowing the assistant to orchestrate among interested agents. Following on from Equation~\ref{eqn:assistant}, we define an evaluation function $\mathbb{E}$ that provides a probability estimate that an agent is going to be able to handle a given event $\mathcal{E}$.

\begin{equation}
\label{eqn:evaluate}
\mathbb{E} : \mathcal{E} \times \Phi \mapsto [0,1]
\end{equation}

Ostensibly, this can be used directly as a signal of whether to invoke an agent or not. For example, an orchestration rule to execute agents that evaluate to a probability above a predefined threshold $\delta$, per turn of user conversation leads to the simplest of aggregated assistants, as described in \cite{rizk2020conversational}.
However, we emphasize once again that agents are sourced and maintained independently, built focusing on their individual tasks. As such, a complex utterance may lower the probability of selection merely due to the fact it is built to receive more focused sentences. That is to say:
\begin{equation}
\mathcal{E}_1 \subseteq \mathcal{E}_2 \centernot\implies 
\mathbb{E}(\mathcal{E}_1, \phi) \leq \mathbb{E}(\mathcal{E}_2, \phi) 
\end{equation}

In the previous example, let's say that there are two separate agents $\phi_f$ and $\phi_h$ for booking flights and hotels, respectively. Then, it is very much possible that evaluating an agent with the entire sentence is less accurate than evaluating a parsed sub-string tailored to those agents: 

\begin{quote}
\em
$S_2$: ``Book me a hotel in New York for this weekend.''
\end{quote}

\begin{quote}
\em
$S_3$: ``Book me a flight from Boston to New York for this weekend.''
\end{quote}

\noindent such that $\mathbb{E}(S_2, \phi_h) \approx \mathbb{E}(S_3, \phi_f) > \mathbb{E}(S_1, \phi_f) \approx \mathbb{E}(S_1, \phi_h)$. This leads us to an intriguing parsing problem: {\em how do we split up the user utterance to facilitate the best expectation calculation for agents in an aggregate?}

In recent work~\cite{clarke2022one}, researchers have attempted to tackle this issue by computing an utterance-to-agent mapping step upfront and an utterance-to-response mapping after all the agents in an aggregate have responded. While their approach and ours operate within an almost identical set of circumstances, we recognize two critical difficulties in building a one-to-one utterance to response classifier: 1) many agents in the aggregate change the state of the world, and hence we cannot depend on the final response to make a routing decision and 2) as described in \cite{aggr}, multiple agents might be combining to service the same utterance, and as discussed in Table \ref{tab:taxonomy} this combination can vary based on the constituents of an aggregate for the same exact utterance. Thus, while an utterance-to-agent match can help in guiding our parse as well (in exchange for an explicit broadcast and evaluation step), our primary aim continues to be to find the optimal parse of an utterance with multiple actors in mind. 

\subsubsection{Decentralized Parsing}
\label{subsubsec:spec}

While this seems like a straightforward parts-of-speech-based parsing problem, a closer inspection reveals a lurking complication. Let us consider a simple syntactic parser based on a context-free grammar to achieve this parse:

{\small
\begin{itemize} 
\item[]
$\mathcal{E} \rightarrow \epsilon$\\[-4ex]
\item[]
$\mathcal{E} \rightarrow$ A\\[-4ex]
\item[]
A $\rightarrow$ ABA\\[-4ex]
\item[]
A $\rightarrow$ CACS\\[-4ex]
\item[]
B $\rightarrow$ then | and | before | after\\[-4ex]
\item[]
C $\rightarrow$ if | when | $\epsilon$\\[-4ex]
\item[]
A $\rightarrow \phi\in\Phi$
\end{itemize}
}

\begin{figure*}[tbp!]
\includegraphics[width=\textwidth]{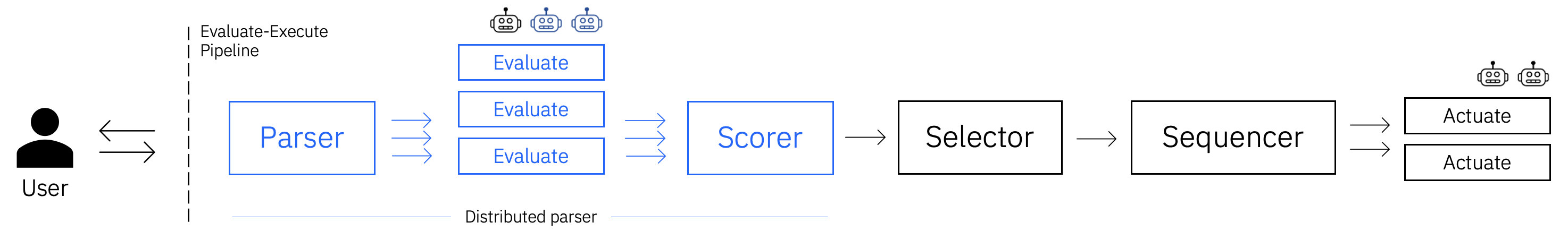}
\caption{
The natural language processing pipeline shows the proposed decentralized parsing algorithm.
}
\label{fig:parser_summary}
\end{figure*}

Interestingly, while this grammar can produce the required splits, it cannot make the final production. Due to the decentralized nature of the knowledge of the system, the correct production can only be determined once $\mathbb{E}(A, \phi)$ has been evaluated $\forall$A. Thus, in order to maximize the effectiveness of the distributed architecture to service a wide variety and complexity of user utterances while preserving agent autonomy, we need a parser that (1) can split up the utterance to maximize multi-intent capture and (2) is able to implement a parsing technique where the evaluation of the goodness of possible splits is done external to the parser.

This is the specification of a parser that can service complex utterances in the aggregated assistant setting. Before we describe how we go about implementing this, it is interesting to note that since the optimal parse itself is dependent on the constituents of an aggregate, the classical notion of multi-intent classification in natural language processing is somewhat muddled here with whether one or more agents are involved in servicing it. 

\begin{table}[h!]
\centering
\begin{tabular}{lll}
\hline
& Single-intent & Multi-intent   \\
\hline
Single-agent & $S_5$: NLQ & -- \\
\hline                        
Multi-agent & $S_4$: NLQ $\rightarrow$ Viz & $S_6$: NLQ, Viz \\
\hline
\end{tabular}
\caption{
Types of multi-intent utterances in the context of a multi-agent conversational system.}
\label{tab:taxonomy}
\end{table}

\paragraph{Multi-Intent Taxonomy}

The simplest case -- single-agent single-intent -- is when an utterance invokes a single agent (Table \ref{tab:taxonomy}). 
Multi-agent single-intent utterances include one intent grammatically, but under the hood, they execute multiple agents due to the way the agents in the aggregate are constructed. Consider the following user utterance: 

\begin{quote}
\em
$S_4$: ``Plot the borrowers' data.''
\end{quote}

The visualization agent (VIZ) first needs to fetch the data to plot, so it has to invoke the information retrieval agent (NLQ) in the background. On the other hand, if the query was simply:

\begin{quote}
\em
$S_5$: ``List all borrower data.''
\end{quote}

\noindent then the execution of the NLQ agent suffices by itself. This shows that even with a traditional single-intent sentence (grammatically), in a distributed parsing problem, we end up with multi-agent splits purely based on the members of the aggregate. In the scope of this work, we focus on multi-agent multi-intent utterances (including the execution of the same agent multiple times), where there are more than two intents explicitly stated in the user utterance -- servicing the single-intent multi-agent cases require higher-order reasoning as demonstrated in \cite{aggr} and is not exclusively a parsing task. 

A multi-intent multi-agent setup covers a wide class of utterances where multiple agents are explicitly invoked with possible orderings and relations among them. One famous class of such utterances are IFTTT (if-this-then-that) instructions\footnote{https://ifttt.com}, where each parameter of the instruction is to be routed to a separate agent, e.g., {\em ``Send me a message when I get an email''} (considering that there are separate messaging and email clients in the assistant). In the running example, this takes the form:

\begin{quote}
\em
$S_6$: ``List all borrower data and plot it.''
\end{quote}

The response from the assistant here should be to execute both NLQ and VIZ agents, but note how with a naive splitting mechanism, such as the one using the syntactic parser in Section \ref{subsubsec:spec}, the system will end up with {\em ``plot it''} as a sub-string that is no good to any agent.

\section{Proposed Framework}

This section describes an implementation of an aggregate assistant architecture that can service complex user utterances based on the specification outlined in Section \ref{subsubsec:spec}. As stated in section \ref{subsec:evaluation}, an evaluation function that calculates an agent's ability to handle the user's request is used to solve the distributed knowledge problem. This forms the basis of an evaluate-and-execute pipeline inside our aggregate agent (Figure~\ref{fig:parser_summary}) with a decentralized parser \tess at the heart of it. The flow of control is conceptualized in Algorithm \ref{algo:algo}.

\begin{figure}[tbp!]
\includegraphics[width=\columnwidth]{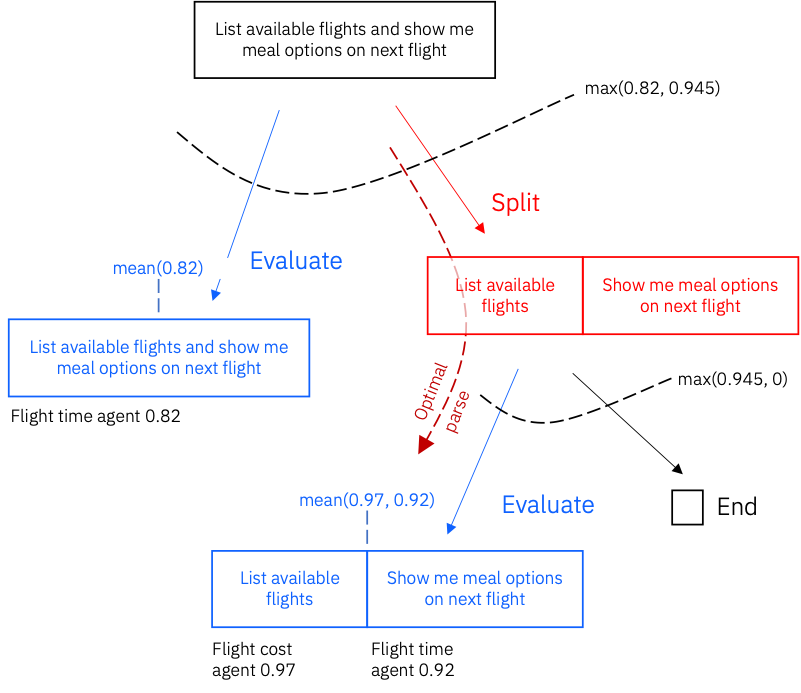}
\caption{
We show here the \tess tree for a simple utterance with one conjunction. Evaluate and parse operations are shown in blue and red, respectively. After the values of the nodes are backed up recursively per Equation \ref{eqn:backup}, the sequence of split-evaluate operations producing the most utility indicates the optimal parse.
}
\label{fig:parse_tree}
\end{figure}

\subsection{The \tess Parser}

The decentralized parser unfolds across three stages in the pipeline. Together, they realize a parse tree, except that the evaluation of the final production rules is done externally. 

\subsubsection{Generating the Parse Tree}

The first step is to generate the \tess tree with all possible parse combinations for the given utterance using specific single-token conjunctions (and, then, after, etc.), double-word conjunctions (first-second, first-after, etc.), punctuation marks (comma, full stop, question mark, exclamation point, etc.), and token-level dependencies (refer to Table~\ref{tab:ex} for examples of correct multi-intent parses).

Each node $\mathcal{N} = \{\mathcal{E}_i\}$ in the parse tree represents the state of the multi-intent parse and is composed of one or more candidate utterances derived from the user utterance. The root is the user utterance itself. The full parse tree is constructed recursively using the following three operations on a node. 

The first one is an \textit{evaluate operation} -- the candidate utterances are
broadcast to the agents to evaluate their likelihood. The score of each utterance in a node is the highest evaluation score of any agent on that utterance, and the evaluation of that node is the average of those individual evaluations.
\begin{equation}
\label{eqn:en}
\mathbb{EN}(\mathcal{N}) = \frac{1}{|\mathcal{N}|} \sum_{\mathcal{E} \in \mathcal{N}} \max_{\phi\in\Phi} \mathbb{E}(\mathcal{E}, \phi)
\end{equation}

\begin{algorithm}[!th]
\SetAlgoLined
\DontPrintSemicolon
\SetKwFunction{FParse}{Parse}
\SetKwFunction{FScorer}{Scorer}
\SetKwFunction{FSelector}{Selector}
\SetKwFunction{FSequencer}{Sequencer}
\SetKwProg{Fn}{Function}{:}{}
\While{$\mathcal{E}$}{
$\{\mathcal{E}\} \leftarrow$ \FParse($\mathcal{E}$)

Find: $\mathbb{E} \subseteq \{\mathcal{E}\}$ 
such that: 

\ForEach{$\Phi \in \{\Phi\}_i$}{
\eIf{$\exists \Phi\textit{.preview}$}{
$P(\Phi) \leftarrow \Phi\textit{.preview}(\mathcal{E})$
\tcp*{confidence}
$S(\Phi) \leftarrow \FScorer(P(\Phi))$
}{
$S(\Phi) \leftarrow 0$
}
}
\vspace{5pt}
$\Phi \leftarrow \FSelector(\Phi)$\;
$\langle\Phi\rangle \leftarrow \FSequencer(\Phi)$\;
\vspace{5pt}
\ForEach{$\phi \in \langle\Phi\rangle$}{
\Return $\phi(\mathcal{E})$
}
}
\vspace{5pt}
\SetKwProg{Pn}{Function}{:}{\KwRet}
\Pn{\FScorer{$P(\Phi)$}}{
\Return $P(\Phi)$
\tcp*{default}
}
\vspace{5pt}
\SetKwProg{Pn}{Function}{:}{\KwRet}
\Pn{\FSelector{$\Phi$}}{
\Return $\Phi$
\tcp*{default}
}
\vspace{5pt}
\SetKwProg{Pn}{Function}{:}{\KwRet}
\Pn{\FSequencer{$\Phi$}}{
\Return {\tt List($\Phi$)}
\tcp*{default}
}
\caption{Flow of control }
\label{algo:algo}
\end{algorithm}

The second one is the \textit{parse} or \textit{split} operation, which implies that we can continue splitting.
\begin{equation}
\mathbb{PN}: \mathcal{N} \rightarrow \mathcal{N}' 
\end{equation}

The last is the \textit{end operation}, which indicates an end to splitting. The score of an end node is 0.

\begin{table*}[tbp!]
\small
\begin{tabularx}{\textwidth}{@{}lXX@{}}
\toprule
\textbf{Parse Type} & \textbf{Utterance} & \textbf{Correct Parse} \\ \midrule
Single-token conjunctions & Book me a hotel and show me flights to NYC  & Book me a hotel + show me flights \\
Double-word conjunctions & First send an email and then a Slack message. & Send an email + Send a Slack message \\
Punctuation marks & Book me a flight, find a cab to the airport & book me a flight + find a cab to the airport \\
Token-level dependencies & Book a hotel and flight to NYC. & Book a hotel to NYC + Book a flight to NYC \\ \bottomrule
\end{tabularx}
\caption{Illustration of different forms of tokens used for multi-intent parsing.}
\label{tab:ex}
\end{table*}

\subsubsection{Evaluation Requests}

After the parse tree is created, the next stage traverses the tree and sets \textit{broadcast scores} for each node. For example, if a node has two candidate utterances derived
from the original utterance ({\em ``list available flights''} and  {\em ``show me meal options for my next flight''}), then each of them will be sent to an agent for evaluation; and for each utterance, we store the agent with the highest confidence along with its confidence score. We show broadcast scores for each evaluate node in Figure \ref{fig:parse_tree}.

\subsubsection{Min-Max Style Backups}

\tess then computes the score of each \textit{evaluate} node as per Equation \ref{eqn:en}. The average corresponds to an average probability mass, but this scoring mechanism can accommodate other schemes like joint or maximum probability. For example, consider the final evaluate operation in Figure \ref{fig:parse_tree}: [``list available flights", ``show me meal options for my next flight"]: {flight-agent: 0.97; meal-agent: 0.92 }; the average probability mass function gives us a selection score of 0.945, whereas the joint probability function gives us a score of 0.892 (probabilities, i.e., utterance evaluations $\mathbb{E}$ are independent).  

At this point, we have a parse tree where each node has a corresponding score. We determine the optimal parsing choice from the root down to the leaves based on a min-max style \cite{aumann1972some} backup computation to compute the score recursively 
$\mathbb{S}$ of each node. 
\begin{equation}
\label{eqn:backup}
\mathbb{S}(\mathcal{N}) = \max \mathbb{EN}(\mathcal{N}), \mathbb{S}\cdot\mathbb{PN}(\mathcal{N})
\end{equation}

\subsection{Selection to Execution}

As described in Algorithm \ref{algo:algo}, once the backups are done, we have an optimal parse of the utterance indicating which agents are involved in acting upon which parts of the user input. This induces a selection of agents to sequence and execute. The order of execution is determined by the sequencer stage and can involve additional reasoning, such as using an AI planner in \cite{aggr}. Since this paper focuses on the parser, we leave the final stages of the pipeline as default, in Algorithm~\ref{algo:algo}, as per \cite{rizk2020conversational}.

\section{Methodology and Evaluation}
\label{exp_methodology}

The first section details the datasets we use to test our framework and baselines. Following that, we discuss the details of NLU model training (i.e., agents). Then, we explain the implementation details of our framework and baselines. We conclude by presenting classification performance and comparison of \tess with baselines. 

\subsection{Datasets}

To test our approach, we refer to popular public datasets for 
intent classification, namely: 
\begin{itemize}[leftmargin=*]
\item 
Single-intent Airline Travel Information System (ATIS) \cite{tur2010left} with 5,871 instances;
\vspace{-5pt}
\item 
Multi-Domain Wizard-of-Oz (MultiWOZ) \cite{budzianowski2018multiwoz, eric2019multiwoz,zang2020multiwoz} with 10,000 instances; 
\vspace{-5pt}
\item 
Multi-intent phrases (MixATIS) \cite{Qin_2020} with 20,000 instances.
\end{itemize}

\paragraph{Survey Dataset}

We also collected an internal dataset where we specifically collected multi-intent multi-agent utterances (according to our taxonomy in Table \ref{tab:taxonomy}) using utterances from ATIS and multiWOZ datasets as references. We asked users to create multi-intent multi-agent utterances considering the given list of agents determined by ATIS and multiWOZ. We collected 60 multi-agent multi-intent utterances for each dataset (for a total of 120 instances) from 6 respondents, 
each providing between 10 and 20 utterances. Two examples are included below, including the multi-intent phrase and the parsed phrase with the intent:

{\small
\begin{itemize}[leftmargin=0pt]
\item[] 
{\em List all the airports in New York. Also, how long will it take me to fly from Arizona to Albuquerque?}
\begin{itemize}[leftmargin=*]
\item List all the airports in New York \textcolor{red}{\tt airport-info}
\item Also, how long will it take me to fly from Arizona to Albuquerque? \textcolor{red}{\tt distance-info}
\end{itemize}
\item[] 
{\em List the cheapest rental cars in Utah and show me the type of aircrafts that are used by Turkish airlines.}
\begin{itemize}[leftmargin=*]
\item List the cheapest rental cars in Utah \textcolor{red}{\tt ground-fare-info}
\item Show me the type of aircrafts that are used by Turkish airlines
\textcolor{red}{\tt aircraft-info}
\end{itemize}
\end{itemize}
}

\subsection{Simulating a Multi-Agent Assistant}
\label{sec:nlu_agent}

\tess is built for a decentralized architecture, 
where each agent is specialized to one specific intent and trained using single-intent sentences. We train a set of NLU models to simulate this setup using the open-source RASA framework \cite{bocklisch2017rasa}. We use pre-trained word embeddings and Dual Intent and Entity Transformer (DIET) classifier \cite{bunk2020diet} to train agents. DIET is a multi-task transformer architecture that simultaneously performs intent classification and entity recognition. It allows plugging pre-trained embeddings such as BERT \cite{devlin2018bert}, ConveRT, or GloVe \cite{pennington2014glove}. More details about this process are available in Appendix \ref{sec:nlu_training}.

\subsection{Baselines}

\paragraph{AGIF} 
The first baseline is  AGIF, a recent multi-intent detection model that also performs slot filling \citet{Qin_2020} -- 
given our multi-intent multi-agent domain, and the outsized importance of slot-filling in conversational agents \cite{louvan2020recent}, this makes for an ideal baseline. We use the open-source implementation\footnote{https://github.com/LooperXX/AGIF} of AGIF (with default parameters). 

\begin{figure}[tbp!]
\centering
\includegraphics[width=\columnwidth]{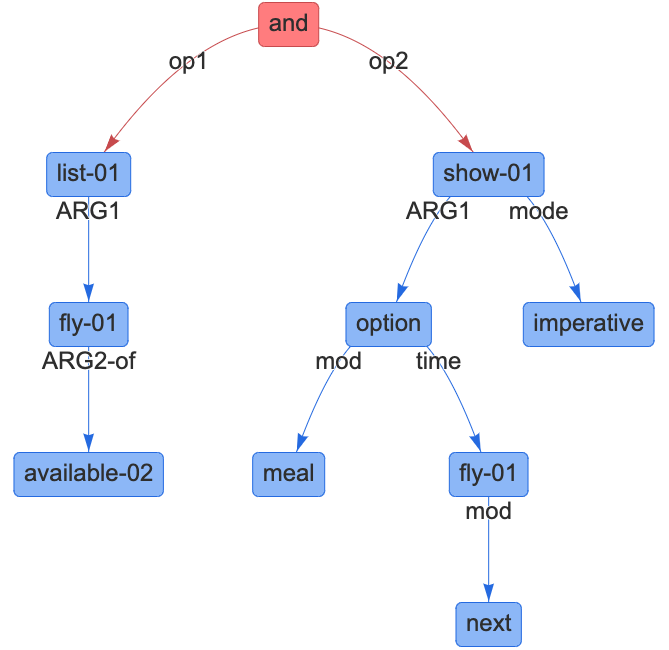}
\caption{
An alternative parse tree using AMR for the utterance: \textit{List available flights and show meal options on next flight}. We assume each sub-tree (under the red arrows) contains a meaningful parse and re-construct parses via traversing the nodes.}
\label{fig:amr_parse_tree}
\end{figure}

\paragraph{AMR}
The second baseline is a parser 
based on Abstract Meaning Representation (AMR)~\cite{banarescu2013abstract}. As a state-of-the-art semantic parser, the AMR baseline gives us an alternative view into the appropriateness or effectiveness of the syntactic parser in \tess for capturing multiple intents in a sentence, which is a common representation in the domain, and it can be used to generate meaningful parses using the semantic information. AMR converts a natural sentence into a rooted directed acyclic graph to capture semantic meaning; concepts and their relationships are captured by nodes and edges respectively~\cite{banarescu2013abstract}. 
We use current state-of-the-art AMR implementation\footnote{https://github.com/IBM/transition-amr-parser}~\cite{zhou2021structure} to obtain a parse for each utterance, which is then used to generate the candidate utterances to evaluate. 

Figure \ref{fig:amr_parse_tree} illustrates the AMR parse tree and tokens we obtain for a specific utterance. After we generate the AMR parse tree, we traverse the tree from the root node in each available direction until each sub-tree (under the red arrows) is completed. Our heuristic approach relies on the idea that each sub-tree represents a meaningful parse. We match tokens with the actual words from the original utterance and re-construct the parsed sentence. For the given example, the generated parses are \textit{list available flights ; show meal options on next flight}. In some utterances, the heuristic approach struggles to find one-to-one mappings with all the available tokens leading to incomplete parses (e.g., \textit{available flights} instead of \textit{list available flights}).

\begin{table*}[tbp!]
\small
\centering
\begin{tabularx}{\textwidth}{X|XXX|XXX}
\hline
& \multicolumn{3}{c|}{Multi-intent ATIS} & \multicolumn{3}{c}{Multi-intent MultiWOZ} \\
\hline
Category & \tess & \tess & AMR  & \tess & \tess & AMR \\
 & (Average) & (Joint) & Parser & (Average) & (Joint) &  Parser\\
\hline
CPCA    & \textcolor{red}{0.34} & \textcolor{blue}{0.33} & \textcolor{olive}{0.02} & \textcolor{red}{0.24} & \textcolor{blue}{0.23} & \textcolor{olive}{0.06} \\

CPWA    & 0.10 & 0.10  & 0.01 & 0.20 & 0.18 & 0.03 \\

WPCA    & \textcolor{red}{0.26} & \textcolor{blue}{0.27}  & \textcolor{olive}{0.48} & \textcolor{red}{0.18} & \textcolor{blue}{0.20} & \textcolor{olive}{0.33}\\

WPWA    & 0.30 & 0.30  & 0.49 & 0.38 & 0.39 & 0.58 \\
\hline
\end{tabularx} 
\caption{
Multi-intent parsing and intent classification accuracy for the survey dataset. Based on correct agent selection accuracy (CPCA + WPCA), \tess (Average or Joint) reaches 60\%, whereas the AMR parser reaches 0.50\%. For the MultiWOZ dataset, \tess (Joint) achieves 43\% whereas the AMR parser achieves 39\%.
}
\label{tab:atis_multiwoz_multiagent_results}
\end{table*}

\subsection{Evaluation Metrics} \label{subsec:evalMet}

To evaluate the classification performance across single-intent sentences, we report the accuracy as the number of correctly identified intents divided by the total number of sentences. For multi-intent sentences, we evaluate the correctness of the selected \textit{parse} as well as the \textit{intent}. In the rest of the paper, we assume that each agent is responsible for a single intent. In other words, the terms ``agent" and ``intent" are used interchangeably. We divide this evaluation into four different categories:
\begin{enumerate}
\item Correct parse and correct agent (CPCA): the selected parse and agent are correct;
\vspace{-5pt}
\item Correct parse and wrong agent (CPWA): the selected parse is correct, but the selected agent is wrong due to the NLU model;
\vspace{-5pt}
\item Wrong parse and correct agent (WPCA): the selected parse is partially correct or incorrect, but the selected agent is correct;
\vspace{-5pt}
\item Wrong parse and wrong agent (WPWA): the selected parse and agent are incorrect.
\end{enumerate}

In this example: {\em ``Give me a list of all airports in Beijing and list the available meals for my next flight''}, there are two ground truth parses: (1) {\em ``give me a list of all airports in Beijing''}, and (2) {\em ``list the available meals for my next flight''}. For the first parse, we expect the label to be the \textit{airport} intent; if our pipeline selects the same intent, it is labeled as the correct agent. To compare parses, we remove punctuation marks and perform (case-independent) character-by-character comparisons.

\subsection{Multi-intent Classification Performance}

The performance of \tess on multi-intent utterances is presented in 
Table \ref{tab:mixatis_multi_intent_results}.
We observe that \tess slightly edges out AGIF in the MixATIS dataset. This could be attributed to mixATIS, including phrases that fit well the heuristic rules adopted in \tess. The accuracy is also influenced by the decentralized intent recognition pipeline from \cite{rizk2020conversational} that is not present in AGIF's pipeline. 

To further understand \tess' performance, we deconstruct its parsing and prediction accuracy into four categories (as described in Section \ref{subsec:evalMet}). Table \ref{tab:atis_multiwoz_multiagent_results} shows that \tess outperforms AMR on every dimension except WPCA. Essentially, even though the AMR parser is generating incorrect parses, the intent recognition component can still correctly identify the users' intents and select the correct agents. We suspect that this has to do with the parsing algorithm creating both fragments of sentences that do not contain enough information to affect intent prediction and those that do contain enough information despite being incomplete parses. However, ultimately, we care about selecting the right agents irrespective of whether the intermediary representation is accurate (i.e., the parses). Hence, looking at CPCA and WPCA, \tess still outperforms AMR (60\% vs. 50\% on ATIS and 43\% vs. 39\% on MultiWoz).  

\begin{table}[h]
\small
\begin{tabular*}{\columnwidth}{l | @{\extracolsep{\fill}} lll}
\hline
Models             & MixATIS    \\
\hline
\tess - Joint     & 0.75         \\
\tess - Average   & 0.76        \\
AGIF               & 0.74          \\
\hline
\end{tabular*} 
\caption{Multi-intent accuracy for MixATIS.}
\label{tab:mixatis_multi_intent_results}
\end{table}

Interestingly, the performance of parsing based on joint versus
average probability is fairly consistent. This can be used as an indicator
of how stable the confidences (determined independently by agents) are --
i.e., if there are over-eager agents in an uncalibrated multi-agent system --
by virtue of how geometric and arithmetic means respond differently 
to outliers \cite{mean}.

Finally, let's consider the computational overhead of adding a parser to the NLU pipeline of chatbots. This is particularly important in applications where computational resources are scarce and conversational systems are not deployed on powerful GPUs for cost reasons. Since \tess does not rely on deep learning approaches (unlike AMR), its parse time is approximately one-tenth of a second, almost 400 times faster than AMR when running on a CPU\footnote{We use 2.3 GHz 8-Core Intel Core i9 CPU with 64 GB 2667 MHz DDR4 memory in all experiments.}, as shown in Table \ref{tab:atis_multiwoz_multiagent_time}. It achieves this without sacrificing prediction accuracy. 

\begin{table}[h!]
\small
\begin{tabular*}{\columnwidth}{l | @{\extracolsep{\fill}} lll}
\hline
Models             & MultiWoz          & ATIS \\
\hline
\tess - Joint     & 0.87               & 0.99  \\
\hline
\tess - Average   & 0.87              & 0.99  \\
\hline
AGIF               & N/A                & 0.97 \\
\hline
AMR Parser         & 0.76             & 0.91 \\
\hline
\end{tabular*} 
\caption{Single intent classification accuracy of \tess and baselines}
\label{tab:single_intent_results}
\end{table}

\begin{table*}[t!]
\centering
\small
\begin{tabularx}{\textwidth}{X|XX|XX}
\hline
         &  ATIS &  & MultiWOZ \\
\hline
Model & Single-intent & Multi-intent & Single-intent & Multi-intent \\
\hline
\tess      & \good{0.06} & \good{0.12} & \good{0.07} & \good{0.1} \\

AMR Parser & 2.86 & 4.34  & 2.80 & 4.81 \\

\hline
\end{tabularx} 
\caption{Average parse time (second) for single-intent and multi-intent sentences in ATIS and MultiWoz datasets.}
\label{tab:atis_multiwoz_multiagent_time}
\end{table*}

Table \ref{tab:single_intent_results} shows that \tess also outperforms the remaining parsers on single-intent phrases. ATIS contained simpler phrases that didn't confuse the parsers (resulting in a high accuracy). In contrast, MultiWoz contained more complex phrases like the example above that resulted in a lower accuracy for \tess compared to the ATIS dataset. Furthermore, errors in the parsing logic of AMR and AGIF caused them to parse some of the single-intent phrases incorrectly, which affected the performance of the downstream task of intent recognition. 

Finally, based on Table \ref{tab:atis_multiwoz_multiagent_results}, \tess does not significantly slow down the response time of the pipeline compared to AMR (especially when GPUs are not available), which is an important consideration for chatbots where responsiveness is crucial. 

\subsection{Single-intent Classification Performance}

Since conversational systems will encounter both multi-intent and single-intent inputs, 
we evaluate the performance of \tess and the other parsers on single-intent phrases. The diversity of natural language phrases may cause parsers to mistake single-intent phrases for multi-intent and parse them accordingly. For example, ``list all borrowers by zip code and yearly income'' is a single-intent phrase despite including the conjunction ``and''. Since these parsers may err, we want to ensure that adding parsers to the NLU pipeline does not deteriorate single-intent recognition performance. 

\section{Related Work}
\label{sec:related_work}

Conversational systems require two main functions to operate: NLU and dialog management. The NLU component understands the user's objective described in natural language by parsing the input (semantically and syntactically), detecting its intent, and identifying entities. This component and its sub-tasks of parsing and intent/entity recognition have seen various approaches over the past few decades, ranging in complexity and efficacy. 

Single-intent classification models have included support vector machines \cite{haffner2003optimizing}, recurrent neural networks \cite{sarikaya2011deep}, and a capsule-based neural network with self-attention~\cite{xia2018zero}. Some models have been jointly trained on the parsing and intent/entity recognition tasks to leverage the correlation between slots and intents~\cite{goo2018slot, liu2019cm, qin-etal-2019-stack}. However, even these models mainly focus on single-intent phrases. 

Multi-intent detection has also been investigated. Perhaps the closest to our work is \citet{kim2017two}, where they proposed a two-stage framework based on a heuristic parser and a classifier. The first stage uses pre-defined conjunctions to parse the sentences and decides whether the sentence is multi-intent based on a heuristic selection algorithm. The second stage was composed of a linear-chain conditional random field trained on multi-intent sentences generated from single-intent sentences. This stage is executed if the first stage cannot detect multiple intents. However, this framework assumed a centralized NLU engine, and their heuristic selection algorithm was unsuitable for our setting with multiple independent agents in the ecosystem. 

Other existing works mainly leveraged an end-to-end deep learning model; \citet{xu2013exploiting} proposed adding features shared by a set of intent combinations and segmenting sentences into word sequences that belong to each single intent. They used log-linear models to perform single-intent and multi-intent detection with their proposed approaches. \citet{gangadharaiah-narayanaswamy-2019-joint} proposed a model composed of LSTMs and performed joint multi-intent classification at sentence-level and token-level. 
To the best of our knowledge, the state-of-the-art was called Adaptive Graph Interactive Framework (AGIF) with an adaptive intent-slot graph interaction layer~\cite{Qin_2020}. This layer was composed of each token’s hidden state generated by a slot-filling decoder and embeddings of predicted multi-intent phrases. Even though end-to-end deep learning frameworks show highly promising results, they are not suitable for decentralized environments, such as in this work, which consists of multiple agents with their own NLU models. 
Additionally, they consume significantly more computational resources than the token-based syntactic parser proposed here. 

\section{Conclusion}
Our new parsing algorithm for scalable, decentralized NLU-based conversational systems matches the performance of deep learning parsers on multi-intent phrases but uses far fewer resources. It shows the potential for efficient non-deep learning solutions without losing accuracy. Future efforts will aim to extend the parser's capabilities to a wider range of multi-intent phrases.

\clearpage

\section*{Acknowledgement}
The authors would like acknowledge the influence of multiple collaborators throughout the various phases of this project that culminated in this paper. Specifically, authors thank Kartik Talamadupula for reviewing the manuscript, Yasaman Khazaeni for her help in developing the problem statement, and the IBM Watson Orchestrate teams who helped mature some of the thoughts around the problem statement through discussions and interactions.

\bibliography{aksar_emnlp22}
\bibliographystyle{acl_natbib}

\clearpage
\appendix
\section{Ethics and Impact Statement}

While usual ethical concerns about conversational agents apply \cite{meyer}, 
an unique concern attached to conversational agents realized
as an aggregate of agents is that of privacy and security \cite{ischen2019privacy}
of user data shared across agents -- this is particularly
of concern since these agents are sourced from different sources
and the assistant by itself may not have full view into the 
inner workings of the agents. 

Additionally, this architectural choice 
raises issues with the transparency of the 
behavior of the assistant, since how the aggregate is 
combining and responding to the user is hidden away from the user.
Some previous works like \cite{aggr} have explored how explainable AI
planning techniques \cite{chakraborti-ijcai-2020} can 
be used to surface those internal behaviors
to engender trust and transparency, as well as enforce 
locking mechanisms on how data is sourced and shared internally.
But there is much work to be done.

On the flip side, this work can have tremendous impact on how the community adopts computationally expensive neural networks. Our work demonstrates that we can outperform deep networks with more computationally efficient approaches (for the sake of the environment and accessibility for less privileged/resource-rich business like small and medium ones).

\section{Limitations}
The space of natural language phrases is large and diverse. The multi-intent heuristic parser we implement improves the intent recognition of a subset of multi-intent phrases: those that fall within the set of sentences defined by the heuristic rules (e.g., include conjunctions). Furthermore, the accuracy of the intent recognition pipeline depends on the accuracy of the decentralized NLU models (which may produce noisy confidences). 
The optimal parse from \tess is only as accurate as the confidences. 
The calibration of confidences coming from independent sources in a multi-agent
system presents a different set of challenges e.g. bandit based approaches have
tried to attempt it in the past \cite{bouneffouf2021toward}. 

\section{Publicly-Available Dataset Details}
We test our framework on the following single-intent and multi-intent datasets. 

\begin{table*}[t!]
  \centering
  \begin{tabular}{lc|lc}
    MultiWoz & & ATIS & \\
    \hline
    Intent & Number of Samples & Intent &  Number of Samples \\
    \hline
    hotel-find & 12244 & flight & 4298 \\
    police-find & 154 & flight-time & 55 \\
    hotel-book & 2111 & airfare & 471 \\
    train-find & 10351 & aircraft & 90 \\
    train-book & 2458 & ground-service & 291 \\
    attraction-find & 4915 & airport & 38 \\
    restaurant-find & 12271 & airline & 195 \\
    restaurant-book & 2526 & distance & 30 \\
    hospital-find & 263 & ground-fare & 25 \\
    - & - & quantity & 54 \\
    - & - & flight-number & 20 \\
    - & - & meal & 12 \\
    
  \end{tabular}
  \caption{We list all intents along with corresponding number of samples for both datasets. We remove some intents from the original datasets to keep the intents same for generating multi-intent multi-agent utterances.}
  \label{tab:1}
\end{table*}

\section{NLU Training}
\label{sec:nlu_training}
\begin{figure*}[t!]
\includegraphics[width=\textwidth]{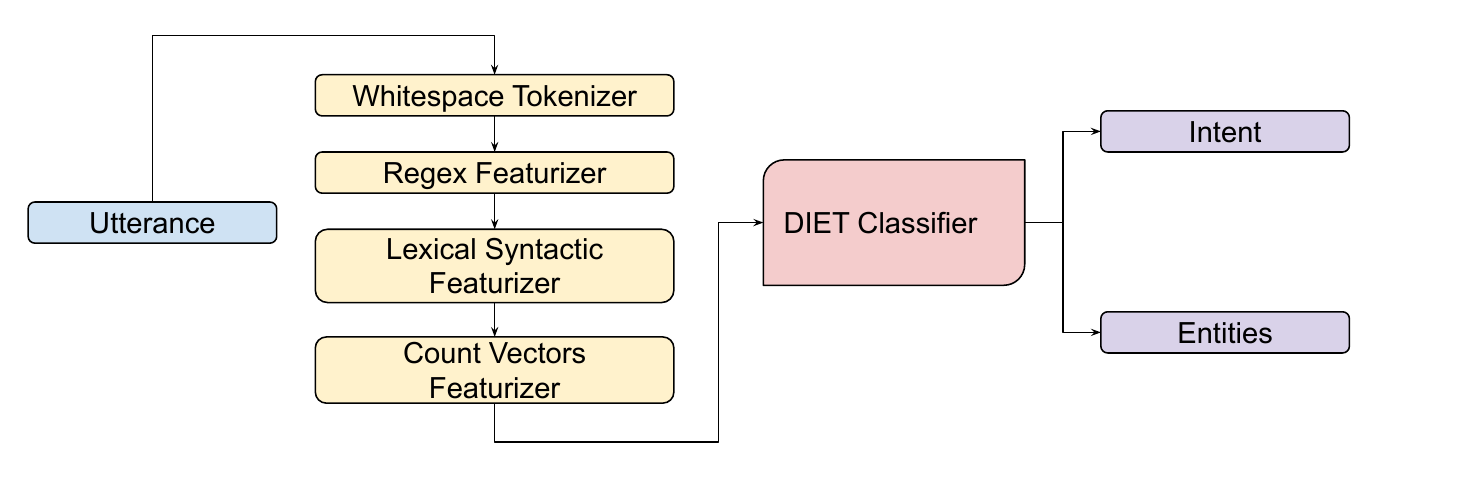}
\caption{Pipeline to train RASA NLU agents. The utterance is processed through different tokenizers and feature extraction methods. Then, we train the DIET model for intent classification and entity detection.}
\label{fig:rasa_diet_flow}
\end{figure*}

The NLU models were trained using the pipeline in Figure \ref{fig:rasa_diet_flow}.

\subsection{Single-intent} Even though we mainly focus on multi-agent multi-intent utterances, our pipeline supports single-intent utterances as well. 
The first dataset is Airline Travel Information System (ATIS)~\cite{tur2010left}, which has 17 different intents, such as \textit{flight} or \textit{meal}. ATIS dataset is composed of 5871 utterances. The semantic frames used to represent the ATIS utterances have slots for phrases and a goal(s) (also known as intent) for each sentence. The values of the slots are not normalized or interpreted.

\begin{figure}[tbp!]
\includegraphics[width=\columnwidth]{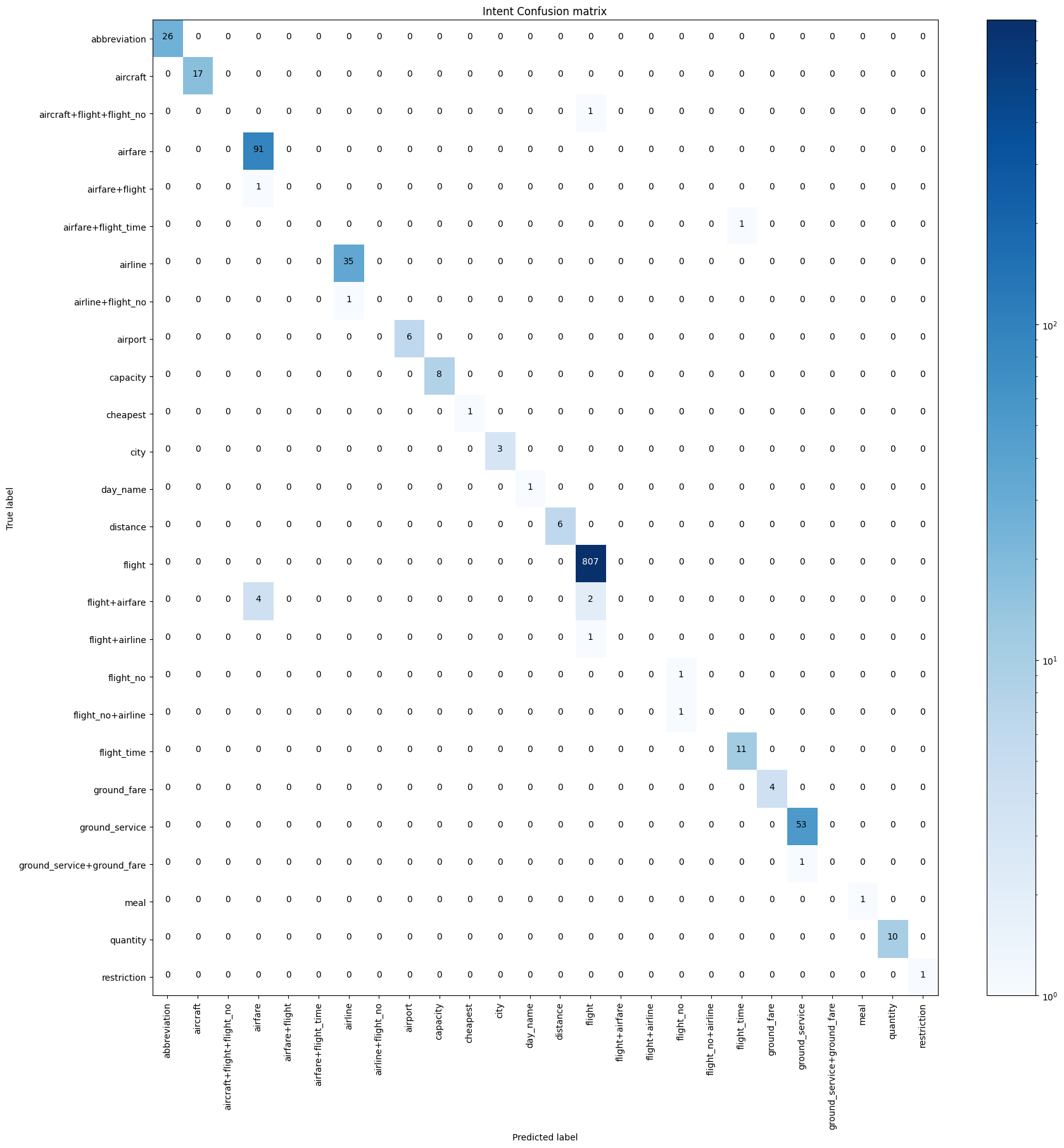}
\caption{Confusion matrix for the trained model using RASA framework in the ATIS dataset}
\label{fig:atis_rasa_cm}
\end{figure}

\begin{figure}[tbp!]
\includegraphics[width=\columnwidth]{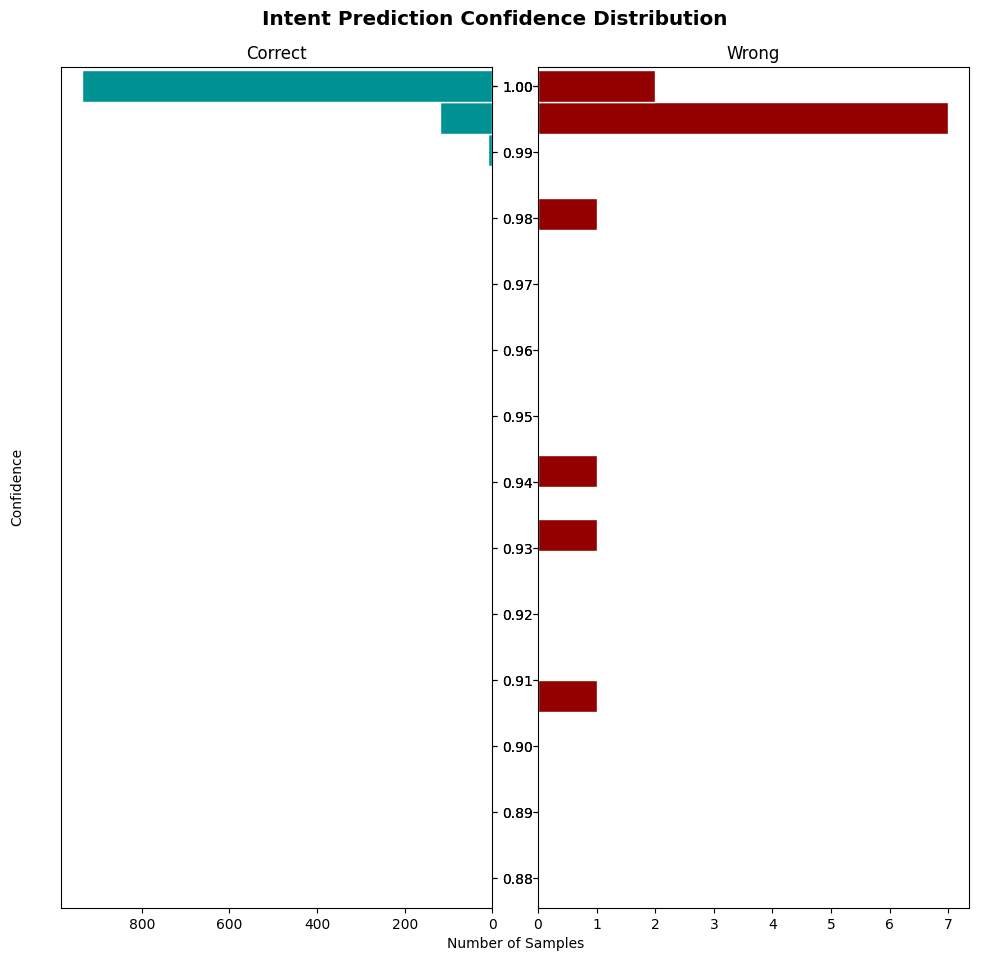}
\caption{Prediction confidences for the samples in the ATIS test dataset}
\label{fig:atis_rasa_hist}
\end{figure}

\begin{figure}[tbp!]
\includegraphics[width=\columnwidth]{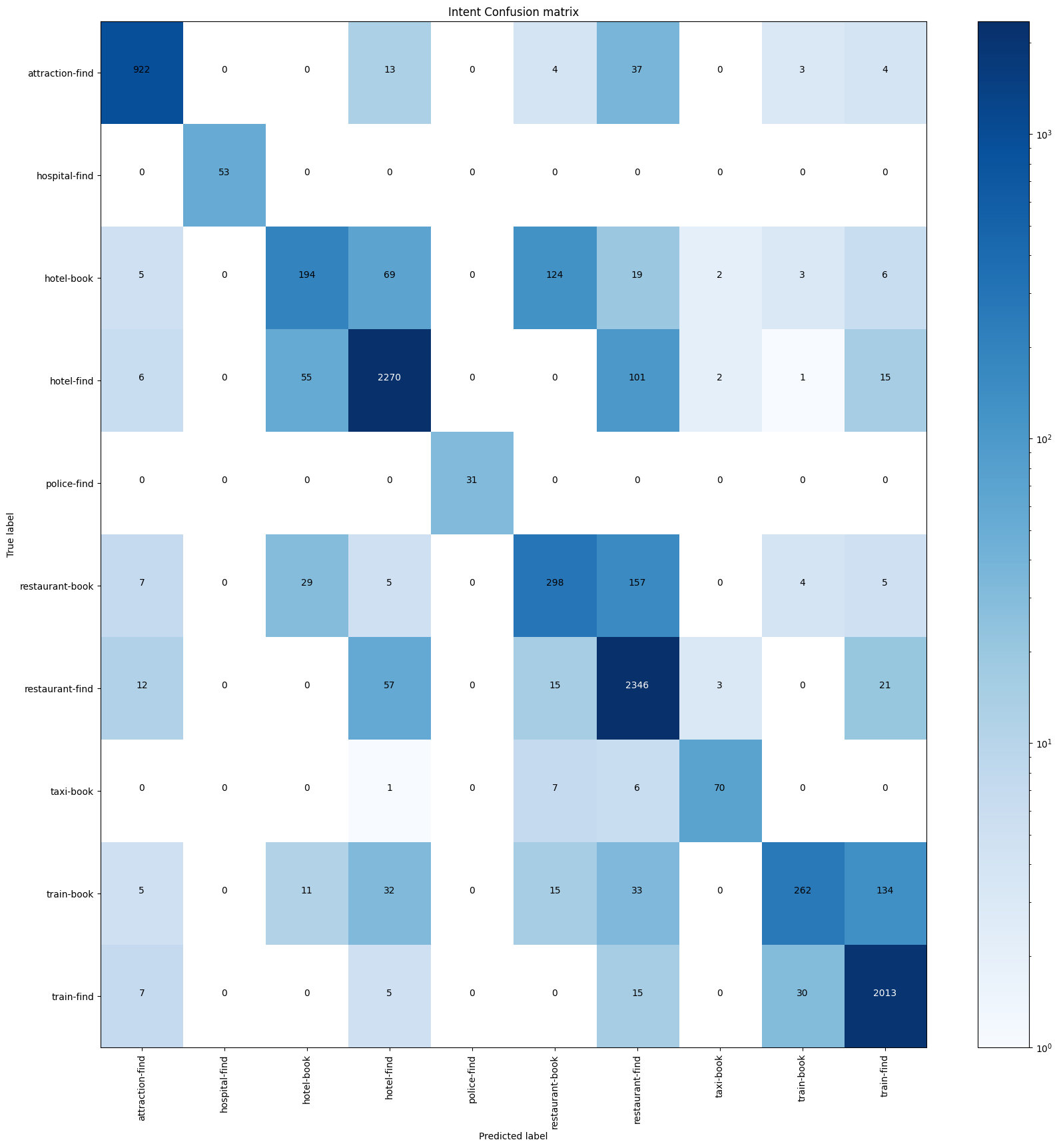}
\caption{Confusion matrix for the trained model using RASA framework in the MultiWoz dataset}
\label{fig:multiwoz_rasa_cm}
\end{figure}

\begin{figure}[tbp!]
\includegraphics[width=\columnwidth]{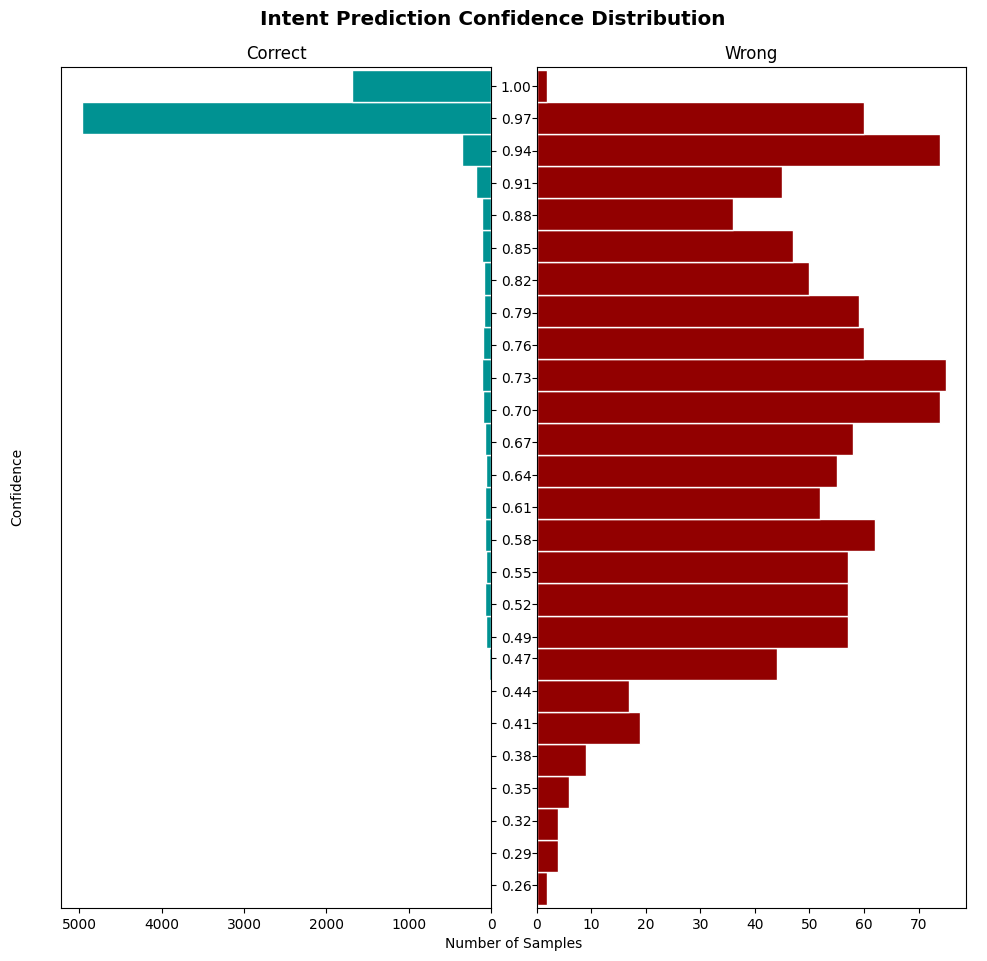}
\caption{Prediction confidences for the samples in the MultiWoz test dataset}
\label{fig:multiwoz_rasa_hist}
\end{figure}

The second dataset is the Multi Domain Wizard-of-Oz dataset (MultiWOZ), which is a large-scale multi-turn conversational corpus with dialogues spanning across several domains and topics~\cite{budzianowski2018multiwoz, eric2019multiwoz,zang2020multiwoz}. In contrast to ATIS dataset, this one includes dialogues from different domains/settings such as \textit{restaurant}, \textit{hotel}, and \textit{taxi} and has about 10,000 dialogues,
which is a lot more than any organized corpus that is currently available. This dataset has 13 different act types such as \textit{inform}, \textit{request}, \textit{book}, and \textit{decline}. \\

\subsection{Multi-intent} In addition to collect our own dataset, we also use the MixATIS dataset~\cite{Qin_2020}. This dataset is composed of multi-intent utterances that are empirically constructed. Authors use some conjunctions (e.g., and, or) and punctuation marks (e.g., comma, semi-colon) to construct the dataset using the intents available in the ATIS dataset. The training set is composed of 18,000 utterances, the validation set is composed of 1,000 utterances and the test set is composed of 1,000 utterances.

\begin{figure*}[t!]
\includegraphics[width=\textwidth]{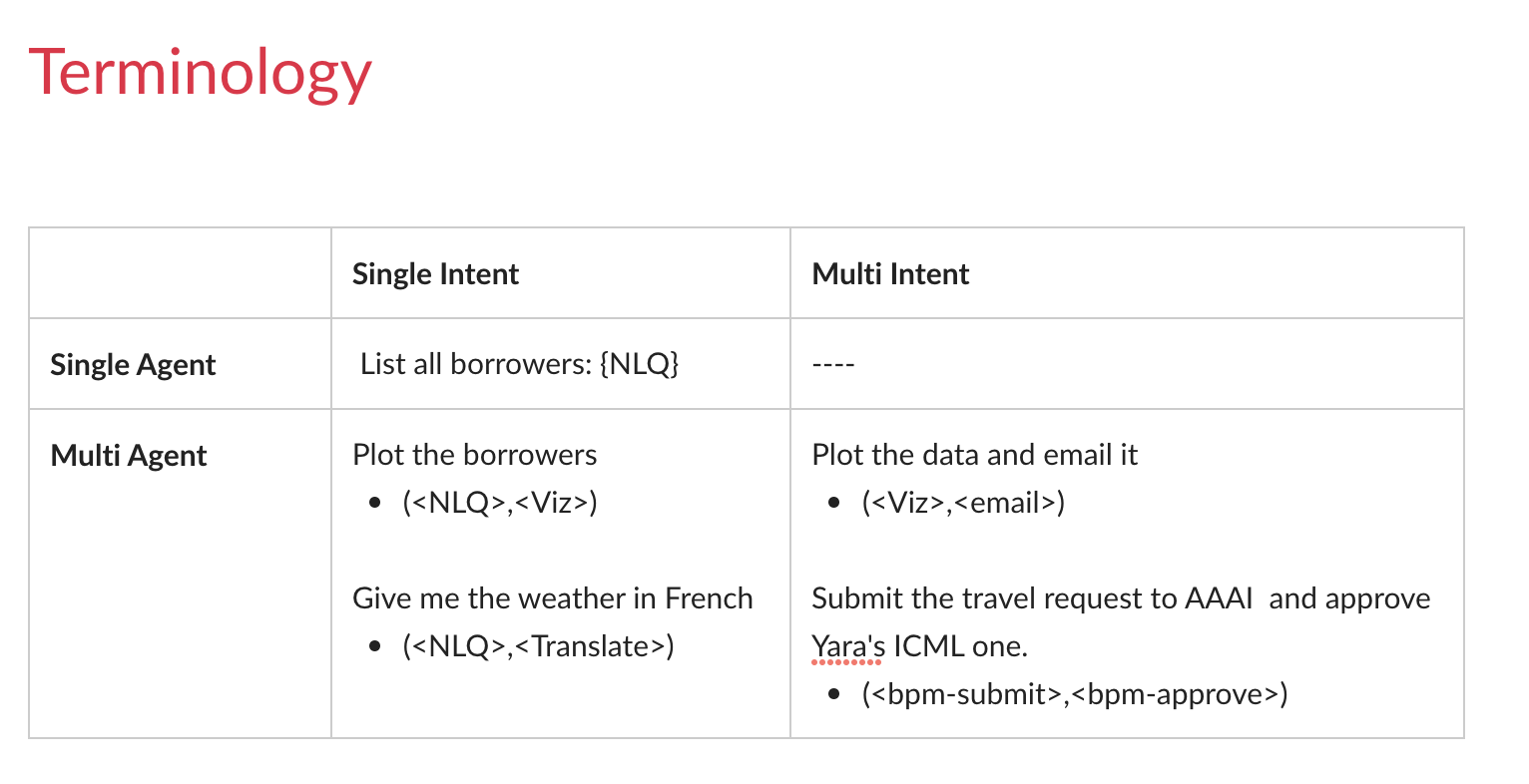}
\caption{Terminology we used for the survey}
\end{figure*}

\begin{figure*}[t!]
\includegraphics[width=\textwidth]{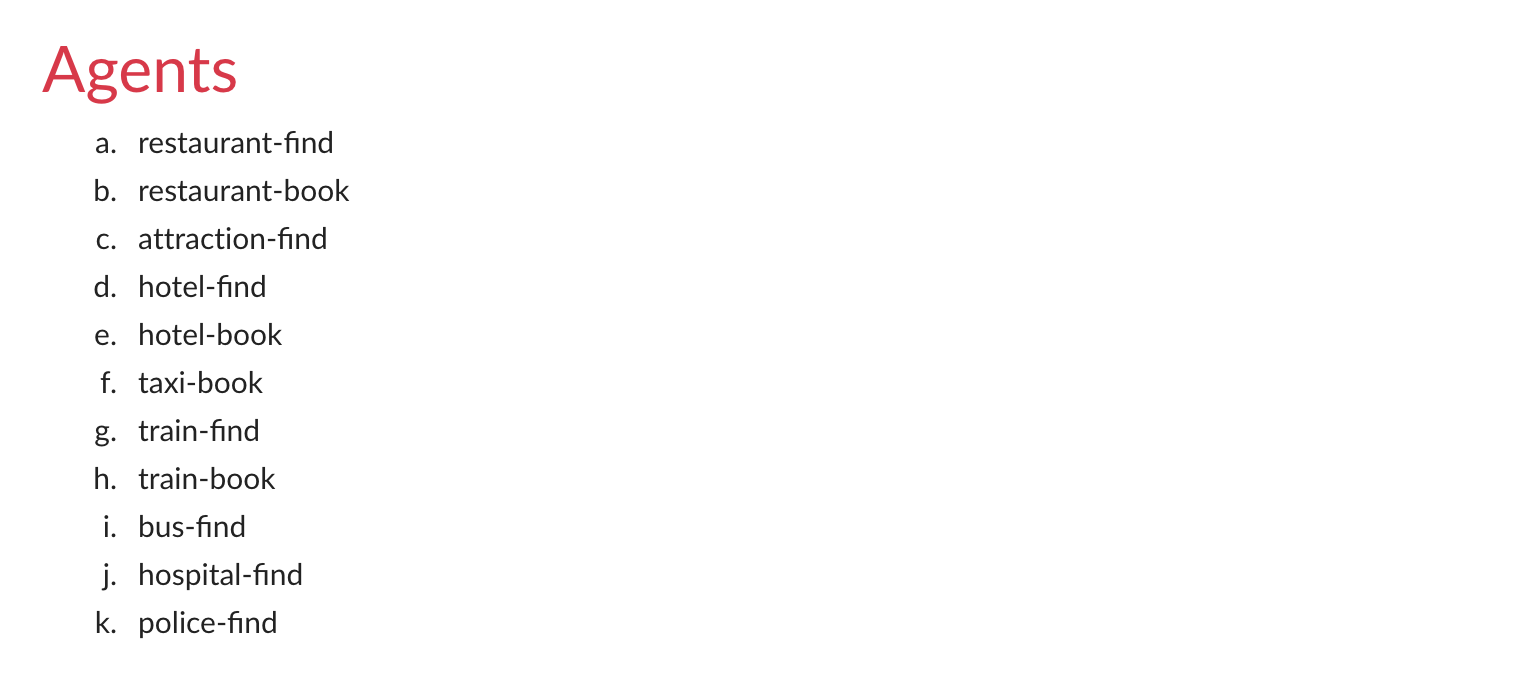}
\caption{MultiWoz dataset agents used in the survey}
\end{figure*}

\begin{figure*}[b!]
\includegraphics[width=\textwidth]{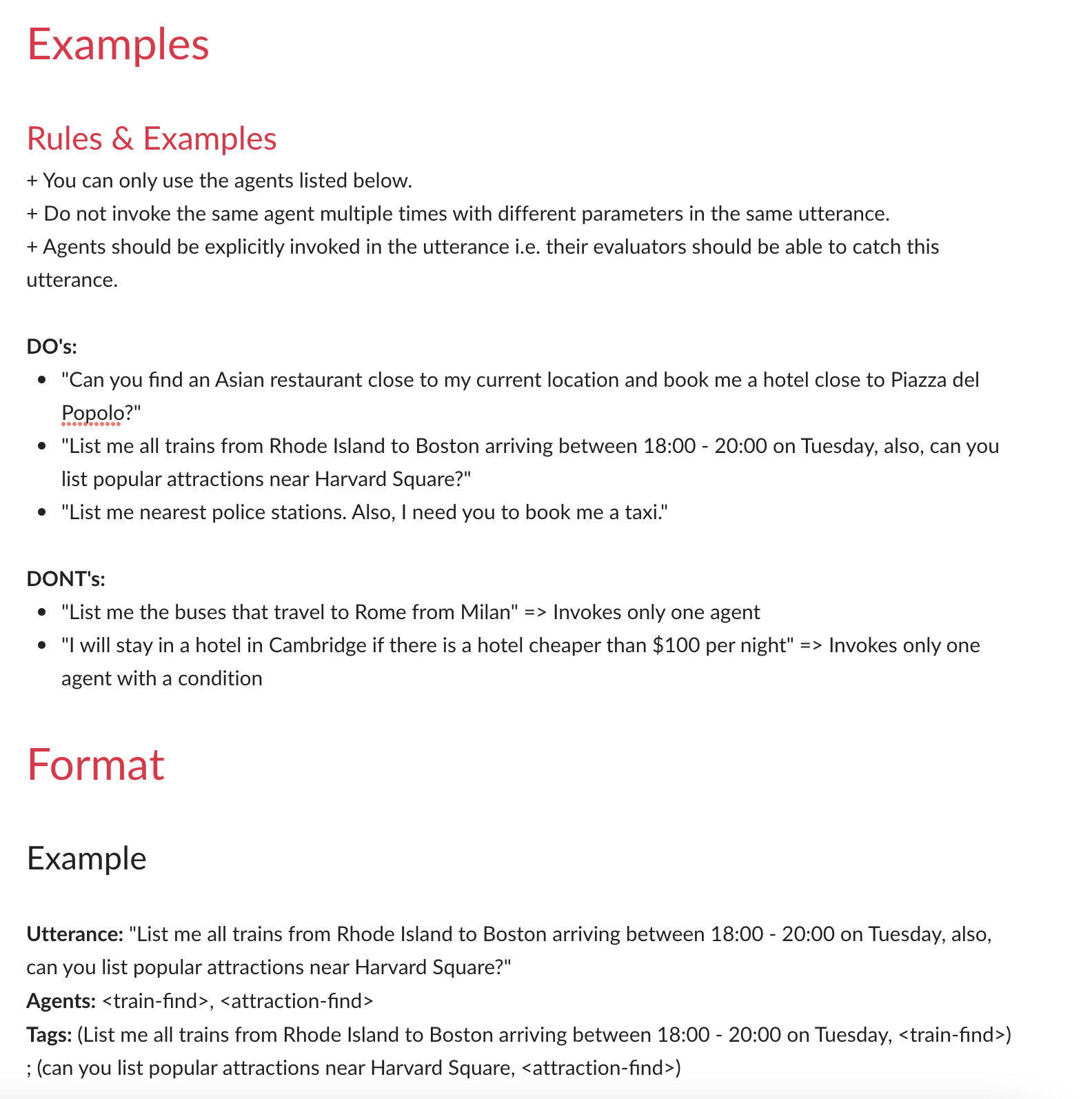}
\caption{Data collection rules and format}
\end{figure*}

\end{document}